\documentclass[sigplan,screen]{acmart}
\usepackage{verbatim}
\usepackage{amsmath}
\usepackage{optidef}

\AtBeginDocument{%
  \providecommand\BibTeX{{%
    \normalfont B\kern-0.5em{\scshape i\kern-0.25em b}\kern-0.8em\TeX}}}

\setcopyright{acmlicensed}

\acmConference[KDD Fashion 2020]{KDD 2020 workshop - AI for Fashion Supply Chain}{August 24--24, 2020}{San Diego, CA}

\begin{document}

\title{Intelligent Warehouse Allocator for Optimal Regional Utilization}

\author{Girish Sathyanarayana}
\email{girish.sathyanarayana@myntra.com}
\affiliation{%
  \institution{Myntra Designs}
  \streetaddress{Bangalore, India}
}
\author{Arun Patro}
\email{arun.patro@myntra.com}
\affiliation{%
  \institution{Myntra Designs}
  \streetaddress{Bangalore, India}
}


\begin{abstract}
In this paper, we describe a novel solution to compute optimal warehouse allocations for fashion inventory. Procured inventory must be optimally allocated to warehouses in proportion to the regional demand around the warehouse. This will ensure that demand is fulfilled by the nearest warehouse thereby minimizing the delivery logistics cost and delivery times. These are key metrics to drive profitability and customer experience respectively. Warehouses have capacity constraints and allocations must minimize inter warehouse redistribution cost of the inventory. This leads to maximum Regional Utilization (RU). We use machine learning and optimization methods to build an efficient solution to this warehouse allocation problem. We use machine learning models to estimate the geographical split of the demand for every product. We use Integer Programming methods to compute the optimal feasible warehouse allocations considering the capacity constraints. We conduct a back-testing by using this solution and validate the efficiency of this model by demonstrating a significant uptick in two key metrics Regional Utilization (RU) and Percentage Two-day-delivery (2DD). We use this process to intelligently create purchase orders with warehouse assignments for Myntra, a leading online fashion retailer.
\end{abstract}

\keywords{Fashion E-commerce; Inventory Allocation; Regional Utilisation; Two-Day-Delivery, Warehouse Assignment; Stock Keeping Unit (SKU); Integer Programming}



\maketitle

\section{Introduction}
Buying and Replenishment is a major, throughout the year activity in large fashion e-commerce firms. Fast changing trends, seasonality, and just-in-time inventory models make inventory procurement an everyday activity. Hundreds of Purchase Orders (POs) are raised every day to procure styles of different article types from vendors. Once the buying quantity is fixed, typically using a demand forecasting model, the next important task is to distribute the procured inventory optimally between the principal warehouses. Logistics cost is a significant cost component in e-commerce operations and reduction in this by a few percentage points will result in cost savings of the order of millions of dollars. This will drive companies towards profitability. Optimal warehouse stocking leads to lower logistics costs and faster delivery of products and hence improved customer experience. An intelligent solution, which can optimally distribute the inventory between warehouses, is of great value.  

The distribution must mirror the geographical demand split. The quantity of the product (SKU) that gets located in a warehouse must be in proportion to the regional demand of the product. This is the localised demand from the areas in the vicinity of the warehouse. Every postal pincode is mapped to the nearest warehouse. Sets of pincodes mapped to the same warehouse partition the whole region into geographical clusters with an associated unique nearest warehouse. The quantity that gets located in a given warehouse must be in proportion to the demand of the associated cluster. The problem is to determine optimal stocking levels for these warehouses. 

Warehouses have capacity constraints and old inventory which must be accounted while determining the new stocking quantities. Geographical distribution of the demand can be different for every SKU. 

\section{Related Work}
Optimal allocation of resources is a well-studied problem across e-commerce and supply chain management. There is literature \cite{stadtler2005supply} for solving problems in production planning, distribution planning, and transport planning using optimization methods. Problems in logistics are modelled using costs such as inventory holding cost, transport cost, etc, and have realistic fixed limits to the variables. \cite{almeder2009simulation} have modelled Supply Chain Networks using a general framework to support the operational decisions using a combination of an optimization model and discrete-event simulation. 

Although some variables are linearly relaxed, the addition of integer constraints like knapsack constraints or capacity in volume or number makes the problem partly discrete and hence needs Mixed Integer Programming methods. 

There has been work using Integer Programming methods to allocate resources optimally between Brick and Mortar and online stores \cite{bhatnagar2014allocating}. \cite{lejeune2008integer} discuss in detail about a production and distribution problem with boolean and general integer variables, formulated as an Integer Linear Program. \cite{lee2008mixed} study an inventory planning problem where the procurement of items and their quantity is modelled using a mixed 0/1 integer program.

\section{Problem Statement and Solution Outline}
The objective is to determine the optimal stocking quantity of each SKU for individual warehouses which will maximise regional utilization. Regional Utilization (RU) is defined as the fraction of total demand which is fulfilled from the nearest warehouse. In cases when warehouses cannot inward the recommended quantities due to capacity constraints, the next nearest warehouse must be prioritised to minimise the cost of redistribution.

We formalise these requirements by defining a penalty matrix $L$ for redistribution costs and expressing the requirements as a constrained optimization problem. Table \ref{tab:notation} lists down all the variables and notation used in the paper. We can formally state the problem as follows.

\subsection{Problem Statement}
Given a Purchase Order (PO) specification: $\{(sku_i , N_i) | 1\leq i \leq M\}$ which is just a set of $(sku_i, quantity)$ pairs for $M$ unique SKUs and $K$ warehouses with capacities $[C_1, C_2,...C_K]$ and existing inventory of the SKUs in warehouses $\{E_{ij}|  1\leq i \leq M; 1 \leq j \leq K\}$, compute the optimal warehouse allocations $\{(sku_i, [X_{i1},...X_{iK},X_{il}]) |1\leq i \leq M\}$ where $X_{ij}$ is the number of $sku_i$ allocated to warehouse $j$. $X_{il}$ denotes the number of $sku_i$ not assigned to any warehouse. This is optionally allowed to handle cases where total SKU quantity is more than the combined capacity of all warehouses. Allocations must satisfy the following constraints.
\begin{align}
    (\sum_{j=1}^{K}X_{ij}) + X_{il} &= N_i && 1 \leq i \leq M\\
    \sum_{i=1}^{M}X_{ij} &\leq C_j && 1 \leq j \leq K
\end{align}

\subsection{Solution Outline}
The solution comprises 2 steps.
\begin{enumerate}
    \item {\textbf{Ideal Splits Computation:} Predicting the ideal split at the SKU level by a split prediction model and generating an ideal warehouse allocation given the order quantity from a demand model for every SKU assuming unlimited capacity in warehouses.}
    \item {\textbf{Optimal Feasible Allocations:} Finding optimal feasible allocations considering warehouse capacity constraints by defining an allocation penalty matrix and solving the constrained optimization problem.}
\end{enumerate}
In the following sections, we will describe the above 2 steps and results in detail.

\begin{table}[]
\caption{Notation of Variables and Constants in the paper}
\label{tab:notation}
\begin{tabular}{l|p{7cm}}
$PO$ & Purchase Order Specification of size $M$ $\{(sku_i, N_i)| 1\leq i \leq M\}$\\
$PO_{i}$ & A tuple of $(sku_i, N_i)$ \\
$N_i$ & Quantity of $sku_i$ to be inwarded \\
$M$ & The total number of unique SKUs in the $PO$ \\
$K$ & Number of Warehouses \\ 
$PO'$ & Exploded Purchase Order Specification of size $N$ $\{(item_i)| 1\leq i \leq N\}$\\
$N$ & Total quantity of all SKUs in $PO$ \\
$E$ & Existing Inventory Matrix $\in Z_{+}^{(M*K)}$\\
$E_{ij}$ &  The existing quantity of $sku_{i}$ in warehouse $j$ \\
$C$ & Warehouse Capacity Vector $\in Z_{+}^{K}$\\
$C_j$ & Capacity of warehouse $j$\\
$X$ & Optimal Warehouse Allocation Matrix  $\in Z_{+}^{(M*(K+1))}$ \\
$X_{ij}$ & The quantity of $sku_i$ to be inwarded to warehouse $j$ for $1 \leq j \leq K$\\
$X_{il}$ & The quantity of $sku_i$ not assigned to any warehouse \\
$P$ & Ideal Split Probability Matrix $\in [0, 1]^{(M*K)}$ \\
$P_{ij}$ & The probability that the purchase event for $sku_i$ is located in geographic region for which nearest warehouse is $j$ \\
$I$ & Ideal Split Matrix $\in Z_{+}^{(M*K))}$ \\
$I_{ij}$ & The ideal quantity of $sku_i$ to be inwarded to warehouse $j$ \\
$Y$ & Binary Decision Variable Matrix $\in \{0,1\}^{(N*K)}$ (formulation 1)\\
$Y_{ij}$ & 1 if $item_i$ is inwarded to warehouse $j$ else 0 (formulation 1)\\
$L$ & Redistribution Cost Penalty Matrix  $\in R^{(K*K+1)}$ \\ 
$L_{ij}$ & The penalty for placing an item which is ideally assigned to warehouse $i$ in warehouse $j$  for $ 1 \leq i, j \leq K$ \\
$L'$ & Truncated Penalty Matrix $L' = L[1:K][1:K] \in R^{(K*K)}$ \\
$\mathbf{1_m}$ & A vector of 1s $\in \{1\}^{(m, 1)}$\\
\end{tabular}
\end{table}

\section{Ideal Splits Computation}
\label{idealsplitscomputation}
Given a purchase order specification: $\{(sku_i , N_i)| 1 \leq i \leq M\}$ and $K$ warehouses, we compute ideal splits $\{I_{ij} | 1 \leq j \leq K\}$ among $K$ warehouses for each $sku_i$: 
\begin{align}
\sum_{j=1}^{K}I_{ij} &= N_i   && 1\leq i \leq M
\end{align}

We assume unlimited capacity in warehouses. Ideal splits $I_{ij}$ are computed by estimating the split probabilities \\ $\{P_{ij} | 1\leq j \leq K\}$ for each $sku_i$ defined as follows.
\begin{align*}
    P_{ij} &= Probability(Purchase\,event\,for\,sku_i\,is\, located\, in\, \\ & geographical\, cluster\, with\, associated\, nearest\, warehouse\,\\ & being \,warehouse j) \label{eqn:ideal_split_probabilities}
\end{align*}

\begin{align}
\sum_{j=1}^{K}P_{ij} &= 1.0
\end{align}

We estimate these probabilities by learning a classifier that predicts the warehouse probabilities using predictors like attributes of the style like colour, brand, fabric, size, age group, price, sleeve length, neck-type, etc. We use around 10 - 12 relevant attributes for each article type and gender group. 

Every purchase event on the platform is associated with a postal pin code that is mapped to the nearest warehouse. This is used to create labelled training records on which the classifier is trained. We tried different classifiers like logistic regression, tree-based classifiers and feed-forward neural network classifiers. Table \ref{tab:classifier} compares the performance of these classifiers. The neural network model outperforms other classifiers and we use a 3-layer MLP for each article type, gender combination in our application. Table \ref{tab:nnlosses} lists the log loss for these classifiers. 

\begin{table}
  \caption{Ideal Split Classifier Log-loss Comparison}
  \label{tab:classifier}
  \begin{tabular}{ccc}
    \toprule
    Classifier&Men's T-shirts&Dresses\\
    \midrule
    Baseline 3-month mean estimate&1.395&1.474\\
    Logistic Regression&1.356&1.392\\
    Random Forest&1.283&1.317\\
    Neural Network&1.040&1.062\\
  \bottomrule
\end{tabular}
\end{table}

\begin{table}[]
    \begin{tabular}{ccc}
    \toprule
    Article Type & Gender & Log Loss \\
    \midrule
    Sweaters  &   Women  &  0.660428 \\
     Jackets  &   Women  &  0.701047 \\
     Jackets  &     Men  &  0.728329 \\
    Sweaters  &     Men  &  0.781279 \\
     Blazers  &     Men  &  0.855442 \\
       Suits  &     Men  &  0.879383 \\
 Sweatshirts  &   Women  &  0.884996 \\
 Casual Shoes  &   Women  &  0.924080 \\
 Sports Shoes  &   Women  &  0.926811 \\
 Sweatshirts  &     Men  &  0.938906 \\
 Sports Shoes  &     Men  &  0.955780 \\
       Heels  &   Women  &  0.984653 \\
 Casual Shoes  &  Unisex  &  0.986829 \\
       Jeans  &   Women  &  0.990782 \\
    Handbags  &   Women  &  0.991535 \\
    Jeggings  &   Women  &  0.993759 \\
  Trolley Bag  &  Unisex  &  0.999156 \\
   Kurta Sets  &   Women  &  1.000210 \\
  Trackpants  &     Men  &  1.005823 \\
 Formal Shoes  &     Men  &  1.014115 \\
 \bottomrule
    \end{tabular}
    \caption{Log Loss for the three layer multi layer perceptron for various article types and genders}
    \label{tab:nnlosses}
\end{table}

Using split probabilities, ideal split for $sku_i$ is computed using Eqn. \ref{eqn:Ideal split eqn}
\begin{align}
    [I_{i1},I_{i2},...I_{ik}] &= N_i * [P_{i1},P_{i2},...P_{ik}] 
    \label{eqn:Ideal split eqn}
\end{align}

If we have existing inventory for the SKUs, $E_{ij}$ being the quantity of existing inventory for $sku_i$ in warehouse $j$, we compute the ideal splits $I_{ij}$ by solving the following system of equations.
\begin{align}
    totalquantity_i &= \sum_{j=1}^{K}I_{ij} + \sum_{j=1}^{K}E_{ij}\\
    \sum_{j=1}^{K}I_{ij} &= N_i \\
    \frac{I_{ij} + E_{ij}}{totalquantity_i} &= P_{ij}  \qquad \qquad 1 \leq j \leq K - 1
\end{align}

Solving these equations may sometimes result in negative values for ideal splits $I_{ij}$ which is obviously not admissible. In such cases, we can compute ideal splits by solving a constrained optimization problem with constraints ensuring that ideal splits are non-negative. It suffices to say that using the classifier output, we can always compute the ideal splits $I_{ij}$. 

Ideal splits are rounded to the nearest integer and any difference in total quantity arising due to rounding is offset using heuristic based rules. At the end of step 1, we have the ideal warehouse allocation matrix $I$. In the next step, we optimally impose capacity constraints on ideal splits.

\section{Optimal Feasible Allocations}
In this step, we optimally impose warehouse capacity constraints on ideal splits computed in the previous section. Optimality is with regard to the inter warehouse redistribution task. We formalise this by defining a redistribution cost penalty matrix $L \in R^{(K*K+1)}$ where

$L_{ij} (j \neq K + 1)$ is the penalty for placing an item which is ideally assigned to warehouse $i$ to warehouse $j$. We set these penalties to mirror logistics cost involved in fulfilling an order from warehouse $j$ when the nearest warehouse is warehouse $i$. 

Non-assignment is allowed to handle cases where total number of items in the PO is greater than the combined capacity of all warehouses. The $(K+1)^{th}$ column of $L$ represents the non-assigment penalty (Eqn . \ref{eqn:non_assignment}). 
\begin{align}
    L_{i, K+1} &= \lambda_{NA}  && 1\leq i \leq K \label{eqn:non_assignment}
\end{align}

 However, we make sure that assignments are always preferred over non-assignments by setting $\lambda_{NA} >> L_{ij}, 1 \leq i, j \leq K$.

Optimization algorithms will always choose assignment over non-assignment because the non-assignment penalty is greater than any assignment penalty.

The key idea behind these optimization formulations is the following. Warehouse allocation implied by ideal splits need not be always feasible because of warehouse capacity constraints. By defining a redistribution penalty matrix, we can compute an optimal re-arrangement of ideal allocations which will be capacity feasible. We can think of the final allocation as minimum distortion from the ideal splits allocation which is feasible with distortion costs defined by the penalty matrix. 

Having defined the redistribution cost penalty matrix $L$, we came up with two novel formulations of the optimization problem. One is a Binary Integer Programming formulation using item-level decision variables and the other is a standard Integer Programming formulation using $sku$ level decision variables. Both are equivalent from an optimality perspective but differ in computational complexity. In the following subsections, we will describe both the formulations.

\subsection{Binary Integer Programming Formulation}
Given a PO specification $\{(sku_i, N_i) | 1 \leq i \leq M\}$, we can define an exploded PO Specification item set $PO'$ by repeating $sku_i$ $N_i$ number of times represented as $item_{ij}$.
\begin{align}
    PO' = \{item_{i,j(i)} | 1 \leq i \leq M; 1 \leq j \leq N_i\}
\end{align}

$item_{ij}$ are $N_i$ identical SKUs of type $sku_i$. Size of the exploded item set is $N = \sum_{i=1}^{M}N_i$ 

For every item in the exploded PO item set, we have an ideal warehouse assignment from ideal split computation from Step 1 (Sec. \ref{idealsplitscomputation}). Warehouse assignments implied by ideal splits are not unique. Different permutations of assignments leading to the same allocation quantities are possible. Different permutations of this warehouse assignments are all equivalent from the point of view of optimization. We can choose any one of them.

Ideal assignment for exploded PO item set = \\
$\{(item_i , w_i)| 1 \leq i \leq N\}$ where $w_i$ denotes the warehouse assigned to item $item_i$. We can represent different warehouses by K-dimensional one-hot vectors like $w_2 = [0,1,...0]$. If an item is not assigned to any warehouse, its warehouse vector will be a K-dimensional zero vector.

Using these one hot representations, we can define an ideal warehouse assignment
matrix $W \in \{0,1\}^{(N*K)}$. Row $i$ of $W$ i.e. $W[i]$ is the ideal warehouse assignment one hot encoding for $item_i$.

For each $item_i$ we define $K$ binary-decision variables, 
$Y_i = [Y_{i1} ,...Y_{iK}]$ which represent the optimal feasible assignment considering capacity constraints. These are determined by solving the optimization problem. 

This gives us the binary decision variable matrix: \newline $Y \in \{0,1\}^{(N*K)}$ to be determined by optimization. Lets define the truncated penalty matrix $L' = L[1:K][1:K] \in R^{(K*K)}$ by dropping the last column representing the non-assignment penalties.

Let $\mathbf{1_m}$ denote a column vector of all 1s of dimension $(m*1)$. We define the non-assigned vector $Z$ as
\begin{align}
    Z = \mathbf{1_N} - Y\mathbf{1_K}
\end{align}

We determine the optimal feasible assignment $Y_{opt}$ by solving the following optimization problem:
\begin{argmini!}
    {Y}{\underbrace{trace(YL'W^{T})}_{assignment\rm\ loss} + \underbrace{\lambda_{NA}\sum_{i=1}^{N}Z_i}_{non\rm\ assignment\rm\ loss} \label{eqn:bip_objective}}{}
    {}
    \addConstraint{\sum_{j=1}^{K}Y_{ij}}{\leq 1 \qquad }{1 \leq i \leq N \label{eqn:bip_optional_selection}}
    \addConstraint{\sum_{i=1}^{N}Y_{ij}}{\leq C_j \qquad}{1 \leq j \leq K \label{eqn:bip_capacity_constraint}}
\end{argmini!}

Optimization objective (Eqn. \ref{eqn:bip_objective}) represents the total redistribution cost for all SKUs given an ideal allocation $W$ and final assignment $Y$. $(YL'W^{T})_{ii}$ represents the redistribution cost for $item_i|i=1,...N$. There are two constraints. Eqn \ref{eqn:bip_optional_selection} constrains that we choose at most 1 warehouse per item. Eqn. \ref{eqn:bip_capacity_constraint} constrains that total items assigned to warehouse $j$ is not more than the capacity of the warehouse $j$.

Using optimal feasible assignments $Y_{opt}$, we compute optimal feasible warehouse splits for SKUs as follows.\\

The ideal assignments for $sku_i$ is $Y(sku_i)$ given by:
\begin{align}
    Y(sku_i) &= [Y_{opt}[j] | 1 \leq j \leq N \, s.t. \, item_j = sku_i] \\
    Y(sku_i) &\in \{0,1\}^{(N_i*K)} \nonumber\\
    X_{ij} &= \sum_{m=1}^{N_i}Y(sku_i)_{mj} \qquad \qquad 1 \leq j \leq K \\
    X_{il} &= N_i - \sum_{j=1}^{K}X_{ij}  
\end{align}

$\{(sku_i , [X_{i1},...X_{iK},X_{il}]) | 1 \leq i \leq M\}$ are the desired optimal feasible warehouse allocations. 

Binary Integer Programming formulation works well for small size orders. The number of decision variables is $O(N*K)$ where $N$ is the total quantity of items and $K$ is the total number of warehouses. For large sized orders, the number of decision variables will become too large and this will be very cumbersome to solve. In the next subsection, we describe an alternative formulation which can efficiently solve large orders.

\subsection{Integer Programming Formulation} \label{ipformulation}
From Step 1 (Sec. \ref{idealsplitscomputation}), we have computed ideal splits:\\ $\{(sku_i , [I_{i1},I_{i2},...I_{ik}]) | 1 \leq i \leq M\}$. This gives us the ideal split matrix: $I \in Z_{+}^{(M*K)}$ where $I_{ij}$ represents the quantity of $sku_i$ assigned to warehouse $j$. Penalty matrix $L \in R^{(K*K+1)}$ is the same as before.

We define a final split Decision Variable Tensor \\ $Y \in Z_{+}^{(M*K*K+1)}$ as follows: \\

\noindent $Y_{iuv} = number \, of \, sku_i \, ideally \, allocated \,to \,warehouse \,u \,which \newline  \,is \,finally \,placed \,in \,warehouse \,v, 1 \leq i \leq M ; 1 \leq u \leq K; 1 \leq v \leq K+1 (v=K+1 \, implies \, not \, assigned \, to \, any \, warehouse)$ \\

We determine optimal final split tensor $Y_{opt}$ by solving the following integer programming problem. 
\begin{argmini!}
    {Y}{\sum_{i=1}^{M}trace(Y_i * L^T) \label{eqn:ip_objective}}{}
    {}
    \addConstraint{\sum_{m=1}^{K+1}Y_{ijm}}{= I_{ij} \qquad }{1 \leq i \leq M; 1 \leq j \leq K \label{eqn:ip_optional_selection}}
    \addConstraint{\sum_{i=1}^{M}\sum_{j=1}^{K}Y_{ijm}}{\leq C_m \qquad}{1 \leq m \leq K \label{eqn:ip_capacity_constraint}}
\end{argmini!}

\begin{figure}
    \centering
    \includegraphics[width=0.5\textwidth]{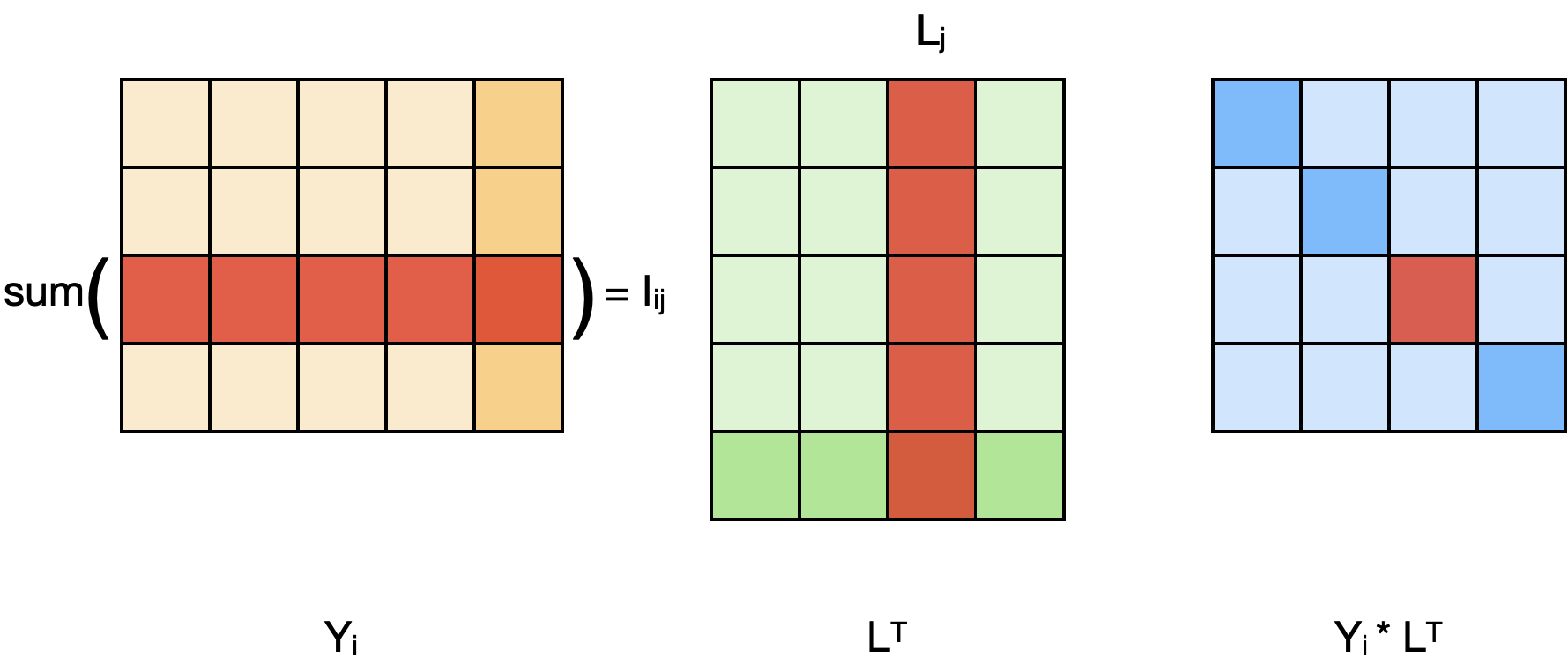}
    \caption{This figure visualizes the matrices. Suppose we have 4 warehouses ($K=4$). $Y_i$ is the decision variable matrix representing warehouse assignment for $sku_i$. $L$ is the penalty matrix. The $(K+1)^{th}$ region is highlighted to represent non-assignment region. $j^{th}$ row of $Y_i$ represents how ideal split $I_{ij}$ gets redistributed between K + 1 warehouses. $j^{th}$ column of $L^T$ represents redistribution losses for a sku ideally bound to warehouse $j$. Therefore, $j^{th}$ diagonal element in the  $Y_i * L^T$ will have the total redistribution loss for items of $sku_i$ which were ideally bound to warehouse $j$.}
    \label{fig:loss_matrix}
\end{figure}

Optimization objective (\ref{eqn:ip_objective}) represents total redistribution cost for $\{sku_i|1 \leq i \leq M\}$ given an ideal split and a final split. The term $trace(Y_i*L^T)$ is the redistribution cost for $sku_i$ as depicted in Figure \ref{fig:loss_matrix}. Eqn \ref{eqn:ip_optional_selection} and  \ref{eqn:ip_capacity_constraint} are optimization constraints for SKU order quantities and  warehouse capacities respectively.

Once we have $Y_{opt}$ by solving the optimization problem, we compute final optimal feasible allocations as follows:
\begin{align}
  X_{ij} &= \sum_{m=1}^{K}(Y_{opt})_{imj} && 1 \leq i \leq M, 1 \leq j \leq K \\
  X_{il} &= \sum_{m=1}^{K}(Y_{opt})_{im,K+1} && 1 \leq i \leq M 
\end{align}

$\{(sku_i , [X_{i1},...X_{iK},X_{il}]) | 1 \leq i \leq M\}$ are the desired optimal feasible warehouse allocations.

The second formulation (Sec. \ref{ipformulation}) is computationally more efficient. The number of decision variables is $O(M*K^2)$ which is usually much smaller than binary integer programming case when $K$ is not very large. With this we can efficiently solve the optimization problem with total SKU quantity ranging in millions, number of SKUs ranging in few thousands and number of warehouses is less than 100 using open source solvers like coin-or/Cbc. 

Note that for this problem, we can always supply an initial feasible solution to warm start the optimization. One trivial feasible solution is the one which does not assign warehouses to any $sku$. This also has the maximum objective value. Using ideal splits and some heuristics, we can supply much better initial feasible solutions to speed up the optimization.

We use this second formulation for our back-testing  and for optimization, we used the coin-or/Cbc solver at \\ https://github.com/coin-or/Cbc. 


\section{Evaluation and Results}
In this section, we describe the back-testing set up and results. We back-tested the model on purchase orders created during an entire month at Myntra. We tested it on major business units like Apparel, Footwear and Personal Care, covering 23 different Article types. Around 8,000 purchase orders were created for 43,000 SKUs and a total quantity of 3.4 million.

We used two warehouse constraint scenarios (Table \ref{tab:constraint 1} and Table \ref{tab:constraint 2}) provided by the Supply-Chain-Management team.

\begin{table}
  \caption{Warehouse capacity constraint - Scenario 1}
  \label{tab:constraint 1}
  \begin{tabular}{ccccccc}
    \toprule
    Month&Type&C1&C2&C3&C4\\
    \midrule
    April 19&Apparel&1005714&502857&377143&754286\\
    &Footwear&91429&45714&34286&68571\\
    &Personal Care&102857&-&17143&-\\
    \midrule
    May 19&Apparel&435810&217905&163429&326857\\
    &Footwear&76191&38095&28571&57143\\
    &Personal Care&85714&-&14286&-\\
  \bottomrule
\end{tabular}
\end{table}

\begin{table}
  \caption{Warehouse capacity constraint - Scenario 2}
  \label{tab:constraint 2}
  \begin{tabular}{ccccccc}
    \toprule
    Month&Type&C1&C2&C3&C4\\
    \midrule
    April 19&Apparel&550979&721502&844438&546241\\
    &Footwear&50089&65591&76768&49658\\
    &Personal Care&56350&-&38384&-\\
    \midrule
    May 19&Apparel&238758&312650&365923&236705\\
    &Footwear&21705&28423&33265&21518\\
    &Personal Care&24418&-&16634&-\\
  \bottomrule
\end{tabular}
\end{table}

We solved the optimization problem using constraints depicted in tables \ref{tab:constraint 1} and \ref{tab:constraint 2}. After every purchase order is created, we deduct the total quantity of the order from the capacities and update the constraints.

Using the optimal splits computed using our optimization algorithm and the purchase data for these skus in subsequent months, we estimated Regional Utilization (RU) which is the fraction of orders which are fulfilled by the nearest warehouse. We also estimated percentage Two-Day-Delivery which is the fraction of orders which are fulfilled within two days of placing the order. Every warehouse has a set of pincodes for which delivery can be done within 2 days. Using this, we can estimate the percentage Two-Day-Delivery. We observed a substantial uptick in these two key metrics --- RU and 2DD over a heuristics based approach previously followed. We compared the estimates with the numbers produced using simple business heuristics. Tables \ref{tab:apparel ru}, \ref{tab:footwear ru} and \ref{tab:pc ru} depict the results.

\begin{table}
  \caption{Apparel RU and 2DD Estimates}
  \label{tab:apparel ru}
  \begin{tabular}{ccc}
    \toprule
    Metric &Scenario 1&Scenario 2\\
    \midrule
    RU (ideal splits)&0.91&0.91\\
    RU (constrained splits)&0.82&0.84\\
    RU (Heuristics based)&0.64&0.64\\
    \midrule
    2DD (ideal splits)&0.64&0.64\\
    2DD (constrained splits)&0.58&0.61\\
    2DD (Heuristics based)&0.48&0.48\\
  \bottomrule
\end{tabular}
\end{table}

\begin{table}
  \caption{Footwear RU and 2DD Estimates}
  \label{tab:footwear ru}
  \begin{tabular}{ccc}
    \toprule
    Metric &Scenario 1&Scenario 2\\
    \midrule
    RU (ideal splits)&0.9&0.9\\
    RU (constrained splits)&0.87&0.9\\
    RU (Heuristics based)&0.38&0.38\\
    \midrule
    2DD (ideal splits)&0.69&0.69\\
    2DD (constrained splits)&0.66&0.69\\
    2DD (Heuristics based)&0.34&0.35\\
  \bottomrule
\end{tabular}
\end{table}

\begin{table}
  \caption{Personal Care RU and 2DD Estimates}
  \label{tab:pc ru}
  \begin{tabular}{ccc}
    \toprule
    Metric &Scenario 1&Scenario 2\\
    \midrule
    RU (ideal splits)&0.93&0.93\\
    RU (constrained splits)&0.7&0.7\\
    RU (Heuristics based)&0.5&0.5\\
    \midrule
    2DD (ideal splits)&0.72&0.72\\
    2DD (constrained splits)&0.53&0.53\\
    2DD (Heuristics based)&0.35&0.35\\
  \bottomrule
\end{tabular}
\end{table}

\begin{itemize}
    \item RU (ideal splits): RU estimate using ideal splits with unlimited warehouse capacities.
    \item RU (constrained splits): RU estimate using optimal feasible splits considering capacity constraints.
    \item RU (Heuristics based): Actual realised RU using a heuristics based warehouse allocation policy.
    \item 2DD (ideal splits): Percentage Two-Day-Delivery estimate using ideal splits with unlimited warehouse capacities.
    \item 2DD (constrained splits): Percentage Two-Day-Delivery estimate using optimal feasible splits considering capacity constraints.
    \item 2DD (Heuristics based): Actual realised Percentage Two-Day-Delivery using a heuristics based warehouse allocation policy.
\end{itemize}

Results clearly show a significant improvement in both RU and 2DD. RU improves by around 27\% over the naive heuristic method and 2DD improves by around 20\%. The performance trend is consistent across all the 3 categories and the 2 constraint scenarios. 

\section{Conclusion}
We have developed an efficient solution to the problem of optimal warehouse allocations from a Regional Utilization perspective using machine learning and optimization methods. We have computed ideal allocations using predictive ML models. We have formalised the requirements of optimal warehouse allocation problem using 2 novel optimization formulations. We have demonstrated the efficacy of the solution with an elaborate back-testing.
We have demonstrated a substantial uptick in key business metrics like Regional Utilization and Two-Day-Delivery over a heuristics based approach.


\bibliographystyle{ACM-Reference-Format}
\bibliography{intelligent_warehouse_allocator}


\begin{thebibliography}{5}


\ifx \showCODEN    \undefined \def \showCODEN     #1{\unskip}     \fi
\ifx \showDOI      \undefined \def \showDOI       #1{#1}\fi
\ifx \showISBNx    \undefined \def \showISBNx     #1{\unskip}     \fi
\ifx \showISBNxiii \undefined \def \showISBNxiii  #1{\unskip}     \fi
\ifx \showISSN     \undefined \def \showISSN      #1{\unskip}     \fi
\ifx \showLCCN     \undefined \def \showLCCN      #1{\unskip}     \fi
\ifx \shownote     \undefined \def \shownote      #1{#1}          \fi
\ifx \showarticletitle \undefined \def \showarticletitle #1{#1}   \fi
\ifx \showURL      \undefined \def \showURL       {\relax}        \fi
\providecommand\bibfield[2]{#2}
\providecommand\bibinfo[2]{#2}
\providecommand\natexlab[1]{#1}
\providecommand\showeprint[2][]{arXiv:#2}

\bibitem[\protect\citeauthoryear{Almeder, Preusser, and Hartl}{Almeder
  et~al\mbox{.}}{2009}]%
        {almeder2009simulation}
\bibfield{author}{\bibinfo{person}{Christian Almeder},
  \bibinfo{person}{Margaretha Preusser}, {and} \bibinfo{person}{Richard~F
  Hartl}.} \bibinfo{year}{2009}\natexlab{}.
\newblock \showarticletitle{Simulation and optimization of supply chains:
  alternative or complementary approaches?}
\newblock \bibinfo{journal}{\emph{OR spectrum}} \bibinfo{volume}{31},
  \bibinfo{number}{1} (\bibinfo{year}{2009}), \bibinfo{pages}{95--119}.
\newblock


\bibitem[\protect\citeauthoryear{Bhatnagar and Syam}{Bhatnagar and
  Syam}{2014}]%
        {bhatnagar2014allocating}
\bibfield{author}{\bibinfo{person}{Amit Bhatnagar} {and}
  \bibinfo{person}{Siddhartha~S Syam}.} \bibinfo{year}{2014}\natexlab{}.
\newblock \showarticletitle{Allocating a hybrid retailer's assortment across
  retail stores: Bricks-and-mortar vs online}.
\newblock \bibinfo{journal}{\emph{Journal of Business Research}}
  \bibinfo{volume}{67}, \bibinfo{number}{6} (\bibinfo{year}{2014}),
  \bibinfo{pages}{1293--1302}.
\newblock


\bibitem[\protect\citeauthoryear{Lee and Kang}{Lee and Kang}{2008}]%
        {lee2008mixed}
\bibfield{author}{\bibinfo{person}{Amy~HI Lee} {and} \bibinfo{person}{He-Yau
  Kang}.} \bibinfo{year}{2008}\natexlab{}.
\newblock \showarticletitle{A mixed 0-1 integer programming for inventory
  model}.
\newblock \bibinfo{journal}{\emph{Kybernetes}} (\bibinfo{year}{2008}).
\newblock


\bibitem[\protect\citeauthoryear{Lejeune and Margot}{Lejeune and
  Margot}{2008}]%
        {lejeune2008integer}
\bibfield{author}{\bibinfo{person}{Miguel~A Lejeune} {and}
  \bibinfo{person}{Fran{\c{c}}ois Margot}.} \bibinfo{year}{2008}\natexlab{}.
\newblock \showarticletitle{Integer programming solution approach for
  inventory-production--distribution problems with direct shipments}.
\newblock \bibinfo{journal}{\emph{International Transactions in Operational
  Research}} \bibinfo{volume}{15}, \bibinfo{number}{3} (\bibinfo{year}{2008}),
  \bibinfo{pages}{259--281}.
\newblock


\bibitem[\protect\citeauthoryear{Stadtler}{Stadtler}{2005}]%
        {stadtler2005supply}
\bibfield{author}{\bibinfo{person}{Hartmut Stadtler}.}
  \bibinfo{year}{2005}\natexlab{}.
\newblock \showarticletitle{Supply chain management and advanced
  planning----basics, overview and challenges}.
\newblock \bibinfo{journal}{\emph{European journal of operational research}}
  \bibinfo{volume}{163}, \bibinfo{number}{3} (\bibinfo{year}{2005}),
  \bibinfo{pages}{575--588}.
\newblock


\end{thebibliography}


\end{document}